\documentclass{article}

    \PassOptionsToPackage{numbers, compress}{natbib}

\usepackage[preprint]{neurips_2026}

\usepackage[utf8]{inputenc} 
\usepackage[T1]{fontenc}    
\usepackage{hyperref}       
\usepackage{url}            
\usepackage{booktabs}       
\usepackage{amsfonts}       
\usepackage{amsmath}        
\usepackage{nicefrac}       
\usepackage{microtype}      
\usepackage{xspace}         
\usepackage{xcolor}         
\usepackage{graphicx}       
\usepackage{multirow}       
\usepackage[textsize=scriptsize,
 disable
  ]{todonotes}      
\usepackage{macros}         
\usepackage{makecell}
\usepackage{listings}
\usepackage[inline]{enumitem}
\usepackage{float}
\lstdefinestyle{prompt}{
  basicstyle=\ttfamily\tiny,
  breaklines=true,
  breakatwhitespace=true,
  breakindent=0pt,
  columns=fullflexible,          
  keepspaces=true,
  showstringspaces=false,
  frame=single,
  xleftmargin=8pt,
  xrightmargin=8pt,
  backgroundcolor=\color{gray!8},
  rulecolor=\color{gray!50},
  framerule=0.4pt,
  tabsize=2,
  literate={-}{{-}}1,           
}

\title{Zero-Shot Goal Recognition with Large Language Models}

\author{%
  Kin Max Piamolini Gusmão\\
  PUCRS\\
  Porto Alegre, Brazil\\
  \texttt{kin.gusmao@edu.pucrs.br}
  \And
  Nathan Gavenski \\
  King's College London\\
  London, England \\
  \texttt{nathan.schneider\_gavenski@kcl.ac.uk} \\
  \AND
  Nir Oren \\
  University of Aberdeen\\
  Aberdeen, Scotland \\
  \texttt{n.oren@abdn.ac.uk} \\
  \And
  Felipe Meneguzzi \\
  University of Aberdeen / PUCRS\\
  Aberdeen, Scotland \\
  \texttt{felipe.meneguzzi@abdn.ac.uk} \\
}

\begin{document}

\maketitle

\begin{abstract}
Large language models have recently reached near-parity with classical planners on well-known planning domains, yet this competence relies on world-knowledge exploitation rather than genuine symbolic reasoning.
Goal recognition is a complementary abductive task structurally better suited to LLM strengths: it consists of evaluating consistency with world knowledge rather than generating novel action sequences.
This paper provides the first systematic zero-shot evaluation of frontier LLMs as goal recognisers on key classical PDDL benchmarks.
Our results show that LLM competence on goal recognition is uneven: some models scale with evidence and approach landmark-based accuracy at full observations, while others remain anchored to world-knowledge priors regardless of how much evidence accumulates.
Qualitative analysis of model reasoning traces reveals that this divergence reflects a fundamental difference in evidence integration rather than domain familiarity.
These findings position goal recognition as a principled benchmark for the foundational planning knowledge of LLMs.
\frm{@Nathan, when I finished writing this I realise this is exactly the opposite of what you suggested, so I will need to work a little on this throughout the day.}
\end{abstract}

\section{Introduction}
\label{sec:intro}
Large language models (LLMs) have recently been adopted as foundational components in a wide range of applications, including those that require reasoning.
Recent benchmarking shows that these models, including reasoning models, have reached near-parity with classical planners on well-known planning domains~\cite{valmeekam2025a,CorreaPereiraSeipp2025a}.
This competence, however, rests on a brittle foundation since performance collapses when domain and object names are obfuscated, revealing that LLMs exploit world knowledge rather than performing genuine symbolic reasoning~\cite{ValmeekamOlmoSreedharanEtAl2022,ValmeekamMarquezSreedharanEtAl2023}.

Goal and plan recognition are related but fundamentally different tasks~\cite{RamirezGeffner2009,RamirezGeffner2010,MirskyKerenGeib2021}.
Given a set of candidate goals and a partial sequence of an agent's observed actions, the task is to infer which goal the agent pursues, alternatively, what plan it uses to achieve such goal.
Unlike planning, goal recognition is abductive: the recogniser does not construct a valid action sequence but instead asks which hypothesis best explains the evidence so far.
Evaluating partial evidence against a fixed hypothesis set aligns naturally with how LLMs leverage the world knowledge encoded in their training, making goal recognition a principled benchmark for probing that foundational planning knowledge.
A model that relies solely on world-knowledge priors will produce stable predictions regardless of how much evidence accumulates, revealing the floor set by domain familiarity; a model that scales with observations demonstrates capabilities beyond that floor.

This paper investigates whether LLM competence on known planning domains transfers to goal recognition.
We run the first systematic zero-shot evaluation of frontier LLMs as goal recognisers on classical PDDL benchmarks, comparing them against the landmark-based approach of \citet{PereiraOrenMeneguzzi2020}.
\frm{I removed the mention of the costs here, but it is a valid argument, in my opinion.}
We evaluate on four domains chosen for their contrasting characteristics (Section~\ref{sec:setup}): \textit{blocks-world}, \textit{campus}, \textit{driverlog}, and \textit{dock-worker robots}.
Our main contributions are:\nl{What would happen if you obfuscate in a slightly different way? Namely take e.g., blocksworld by say you are stacking elevators (or even something like balls as intuitively they would rolll of each other, i.e., you are using LLM domain knowledge against it?}\frm{Good further work.}\frm{Revise this when we have full results.}
\begin{enumerate*}[label=(\roman*)]
  \item The first systematic zero-shot evaluation of frontier LLMs on classical PDDL goal recognition benchmarks, comparing against landmark-based goal recognition~\cite{PereiraOrenMeneguzzi2020}.
  \item Identification of three qualitatively distinct LLM behaviour patterns: \gptfour achieves competitive recall at low observation counts but never improves with more evidence; \gptoss and \gptfive scale with observations, approaching landmark-based accuracy at full observations; and \qwen fails almost entirely on all but the simplest domain.
  \item Evidence that high spread is a shared failure mode for LLMs at sparse observations, and that accumulating evidence is not a solution for all models.
  \item Our results provide a valuable extension to the insights from the mystery-domain findings of \citet{ValmeekamOlmoSreedharanEtAl2022,ValmeekamMarquezSreedharanEtAl2023}. Even on familiar domains with standard names, goal recognition shows that frontier LLM performance deteriorates as plans and the space of hypothesis grow larger, suggesting that the bottleneck is not whether LLMs know the domain but whether they can integrate a growing sequence of observations into a coherent prediction.
  \item A qualitative analysis of model reasoning traces that identifies shared failure modes (confabulation, overconfidence at sparse observations, position bias, and errors at high observability) and derives concrete interventions to guide the development of LLM-based goal recognisers.
\end{enumerate*}
\section{Background}
\label{sec:background}

\label{sec:bg:gr}

Goal recognition is the problem of inferring the goal of an observed agent from a partial sequence of its actions.
This task is intimately related to that of planning, which consists of determining how to achieve a given goal $\goal{}$ from a known initial state $\initialstate{}$. 
However, instead of having to simulate paths from $\initialstate{}$ towards $\goal{}$, goal recognition is an inverse problem, that of inferring the true goal $\realgoal{}$ of the observed agent from observed actions.
\citeauthor{RamirezGeffner2009}~[\citeyear{RamirezGeffner2009,RamirezGeffner2010}] formalise this problem in terms of classical planning, and \citet{MirskyKerenGeib2021} survey its breadth.

\begin{definition}[Goal Recognition Problem]
\label{def:gr-problem}
A \defterm{goal recognition problem} is a tuple $\recognitionproblem = \tuple{\domain, \initialstate, \goals, \observations}$, where $\domain$ is a planning domain, $\initialstate$ is the initial state, $\goals = \{\goal_1, \ldots, \goal_n\}$ is a finite set of candidate goal hypotheses, and $\observations = \tuple{\action_1, \ldots, \action_k}$ is an ordered sequence of observed actions.
The true goal $\realgoal \in \goals$ is unknown.
\end{definition}

Given a goal recognition problem, the task is to return a ranking or probability distribution over $\goals$ given $\observations$, identifying $\realgoal$ as the most likely candidate.
In practice, many approaches return the probability distribution over all $\goal \in \goals$ rather than a strict ordering, and allow for ties.
We note that the notion of \defterm{partial observability} in goal recognition differs from partially observable Markov decision processes~\cite{MeneguzziPereira2021}.
The recogniser never observes the world state directly; it receives only an ordered subset of the actions the agent executes, and must infer the goal from that incomplete trace.

\begin{definition}[Observation]
\label{def:obs-sequence}
Let $\plan = \tuple{\action_1, \ldots, \action_m}$ be the full plan executed by the observed agent.
An \defterm{observation sequence} is a tuple $\observations = \tuple{\action_{i_1}, \ldots, \action_{i_k}}$ where $1 \leq i_1 < i_2 < \cdots < i_k \leq m$, that is, an ordered subsequence of $\plan$ that preserves action order but may omit actions at any position.
The \defterm{observation percentage} is $100 \cdot k / m$.
\end{definition}

Because $\observations$ is a subsequence rather than a prefix, the recogniser has no information about actions at unobserved positions.
Observations consist of ground actions rather than state snapshots; this is the standard assumption in classical goal recognition~\cite{RamirezGeffner2009,MeneguzziPereira2021}.
We assume all observed actions are accurate, i.e., that the partial observation is an ordered subset of $\plan$.
Noisy observations, in which some observed actions are replaced by incorrect ones, form a distinct problem setting that we explicitly exclude.
At low observation percentages the evidence is sparse and many candidate goals remain consistent with $\observations$, and as the recogniser receives more actions, disambiguation becomes easier.

\label{sec:bg:landmarks}

\citet{PereiraOrenMeneguzzi2020} develop an efficient approach to goal recognition based on planning \emph{landmarks}.
A landmark for a goal $\goal$ is an action that any agent pursuing $\goal$ must eventually execute, regardless of which specific plan it follows; the set of all such actions is $\lm{\goal}$.
Computing $\lm{\goal}$ is expensive, in the worst case as hard as planning itself~\cite{HoffmannPorteousSebastia2004}, but it can be approximated in polynomial time via delete-relaxed landmark extraction and done offline once per problem.
The resulting landmark sets encode knowledge about each goal analogously to how an LLM's parameters encode world knowledge, i.e., a costly upfront computation that makes subsequent inference cheap.

Given an observation sequence $\observations$, recognition then proceeds by counting which landmarks each candidate goal has already achieved.
The landmark score of $\goal$ is the fraction of its landmarks observed so far $\lmscore{\goal}{\observations} \;=\;
\nicefrac{\vert\{l \in \lm{\goal} : l \in \observations\}\vert}{\vert\lm{\goal}\vert}$
%
%
The most-likely goal set is $\predgoals = \argmax_{\goal \in \goals}\, \lmscore{\goal}{\observations}$.
Because landmark achievement accumulates monotonically as $\observations$ grows, $\lmscore{\realgoal}{\observations}$ tends to increase relative to false goals with each additional observation, so recognition accuracy improves steadily as the observation percentage rises.
This monotonic scaling behaviour is the key property against which we compare LLM-based recognisers in Section~\ref{sec:results}.

\section{Zero-Shot Goal Recognition with Large Language Models}
\label{sec:method}

A recogniser requires world knowledge of planning domains for zero-shot goal recognition, as well as the ability to update its goal assessments as new observations arrive, a demand that pattern matching alone cannot satisfy.
This section introduces our approach to zero-shot goal recognition using LLMs.
We first describe our prompting strategy, then situate goal recognition within the complexity landscape of planning to motivate it as a natural benchmark for the foundational planning knowledge of LLMs.

\subsection{Prompting Strategy}
\label{sec:method:prompting}

We treat zero-shot goal recognition as a structured prompting task.
We provide no fine-tuning, few-shot examples, or domain-specific context.
The evaluation is strictly zero-shot to isolate the contribution of the LLM's world knowledge.

The prompt has four parts.
The first part assigns the model a role as an expert in AI planning and describes each input it will receive:
\begin{enumerate*}[label=(\roman*)]
    \item the PDDL domain with its action schemas (names, typed parameters, preconditions, and effects);
    \item the problem template with the list of objects, the initial state, and a placeholder for the goal;
    \item the candidate goal set $\goals$, where each $\goal \in \goals$ is a conjunction of facts; and
    \item the observation sequence $\observations$ as an ordered list of grounded actions.
\end{enumerate*}
The second part states the task: for each $\goal \in \goals$, assign a likelihood score in $[0,1]$ such that the scores sum to one, then return the set $\predgoals$ of goals tied at the highest score, together with a free-text explanation of the reasoning.
The third part inserts the concrete problem instance.
The fourth part provides a rigid output format template, listing each hypothesis with its score, the most-likely goal set $\predgoals$, and the reasoning trace, and instructs the model to follow it exactly, without additional text or formatting.
This explicit format constraint allows us to parse the model's output programmatically without post-hoc cleaning.
Figure~\ref{fig:prompt-template} in Appendix~\ref{app:prompt} shows the full template.

The normalised scoring requirement follows directly from the formalisation of goal recognition in Section~\ref{sec:bg:gr}. 
The standard output of a recogniser is a probability distribution over $\goals$~\cite{RamirezGeffner2009,RamirezGeffner2010}, and requiring scores to sum to one operationalises this within the prompt.
The free-text explanation serves one purpose: it provides interpretable evidence of the model's decision process for the qualitative analysis in Section~\ref{sec:discussion}.
Because the output format places the reasoning block \emph{after} the scores and the predicted goal set, the explanation is a post-hoc rationale rather than a chain-of-thought that guides the likelihood assignments~\cite{wei2022chain}.
We adopt the standard correctness criterion of \citet{RamirezGeffner2009,RamirezGeffner2010,PereiraOrenMeneguzzi2020} whereby a recogniser is correct on a problem when the true goal $\realgoal$ is contained in the returned set $\predgoals$. 
This is the same criterion from the landmark-based baseline against which we compare.

\subsection{Goal Recognition as an LLM Benchmark}
\label{sec:method:complexity}

\citet{RamirezGeffner2009,RamirezGeffner2010} formalise goal recognition as a compilation to planning: for each candidate goal $\goal \in \goals$, two planning problems are solved to determine whether the observations $\observations$ are consistent with an optimal plan for $\goal$.
Since classical propositional planning is PSPACE-complete~\cite{Bylander1994} and this compilation preserves that hardness~\cite{Nebel2000}, goal recognition via planning is no easier than planning itself.
\citet{PereiraOrenMeneguzzi2020} avoid the compilation entirely, and separate the complexity of understanding the domain dynamics from that of incorporating observation evidence.
First, their approach extracts landmarks for each goal hypothesis ($\lm{\goal}$).
Then, it scores each candidate goal by comparing observations with the precomputed landmarks.
This yields a complexity of $O(|\goals| \times |\lm{\goal}| \times |\observations|)$, where $|\lm{\goal}|$ is bounded by the number of ground actions in the domain.
LLM-based recognition requires a single forward pass over the prompt, with cost dominated by the size of the input rather than the planning state space.

The key structural distinction is between \emph{generative} and \emph{discriminative} tasks.
Planning asks a model to \emph{produce} a valid action sequence from scratch, requiring consistent long-horizon causal reasoning.
Goal recognition asks a model to \emph{evaluate} whether an observed sequence $\observations$ is consistent with each candidate goal $\goal \in \goals$, a discriminative task.
Given the same PDDL representation, goal recognition never requires generating actions, since the model already observed the agent's behaviour, and the model's role is to interpret it.
This distinction suggests that LLMs should be structurally better suited to goal recognition than to planning, since evaluating consistency with world knowledge is closer to what LLMs reliably do well~\cite{ValmeekamOlmoSreedharanEtAl2022,ValmeekamMarquezSreedharanEtAl2023}.

Beyond its practical interest, goal recognition constitutes a principled benchmark for the foundational planning knowledge encoded in LLMs.
A model with strong domain knowledge will have prior beliefs about which goals are likely that suffice at low observability, where little evidence is available to discriminate between hypotheses.
A model relying solely on such priors, however, will produce approximately constant predictions regardless of how much evidence accumulates; correct recognition at higher observability additionally requires updating those beliefs incrementally as each new action is observed, a capability that world knowledge alone cannot provide.
We therefore hypothesise that LLMs will perform competitively against landmark-based recognition at low observation rates, where foundational knowledge priors suffice, but that only models capable of integrating new evidence will improve further.
The resulting divergence (a world-knowledge floor at sparse observations and an evidence-dependent ceiling thereafter) is the central signal our experiments are designed to measure, and is what distinguishes world-knowledge retrieval from genuine evidential reasoning.

\section{Experimental Setup}
\label{sec:setup}

To understand the capabilities of LLMs in solving goal recognition problems, we compare them against a state-of-the-art symbolic approach, namely the landmark-based method by~\citet{PereiraOrenMeneguzzi2020}.
For the LLMs, we select two commercial and two open-weight models. 
Among commercial models, we choose OpenAI's \gptfour, for offering a balanced trade-off between cost and performance, and \gptfive, the frontier OpenAI model at the time of our experiments, for representing the upper bound of available proprietary capability.
Among open-weight models, we choose \gptoss-20B, the open-weight variant of the GPT line\footnote{\href{https://openai.com/index/introducing-gpt-oss/}{OpenAI's GPT-OSS}; available via \href{https://ollama.com/library/gpt-oss}{Ollama}.}, enabling a direct comparison between open and closed models within the same training lineage, and \qwen-9B\footnote{\href{https://qwen.ai/blog?id=qwen3.5}{Alibaba's Qwen 3.5-9B}; available via \href{https://ollama.com/library/qwen3.5:9b}{Ollama}.}, for representing a competitive open-weight model from a distinct model family and organisation, deployable locally without API dependency.
We run the experiments on a machine equipped with a 2GHz Intel Xeon Gold 6138 CPU with 32 GB of RAM and an NVIDIA RTX 2080 Ti GPU with 11 GB of VRAM.

%
%

We draw problems from the goal and plan recognition benchmark of \citet{PereiraMeneguzzi2017}, which comprises $15$ PDDL domains.
Due to the cost of running commercial LLM APIs at scale, we evaluate on four domains due to their varied characteristics: \emph{Blocks World}, \emph{Campus}, \emph{Driverlog}, and \emph{Dock-Worker Robots} (\emph{DWR}).
\emph{Blocks World} is a classic combinatorial domain with a vocabulary well-represented in LLM training data; failure here cannot be attributed to an unfamiliar vocabulary, making it the primary test of evidence-integration ability.
\emph{Campus} is a simple control domain with short plans and a small goal space; it serves as a minimum competence check, since any approach that fails here is operationally useless.
\emph{Driverlog} is a moderately difficult domain with medium-length plans, a large goal space, and real-world logistics vocabulary (drivers, trucks, packages, locations), testing whether LLM world knowledge transfers to a combinatorially harder setting.
\emph{DWR} is a difficult domain for both classical search and LLM-based recognisers, with very long plans, large goal spaces, and many interdependent objects to track across steps.

For each domain, the problems are divided into five observability degrees: $[10\%, 100\%]$, in increments of $20\%$, of the plan length.
Recall that we define the observability in goal recognition as the percentage of the plan's actions that are observed by the recogniser.
Therefore, the lower the observability degree, the harder the problem is as more goals may share the observed plan.
\citet{PereiraMeneguzzi2017} generate each problem by randomly retaining actions from the full plan with probability equal to the observability degree, running this sampling up to three times per plan and discarding duplicate observation sets; at $100\%$ observability the recogniser always receives the full plan, so each plan yields exactly one problem, resulting in fewer problems at full observability than at partial observability.
The domains also vary in the average number of goal hypotheses per problem and the average number of landmarks per hypothesis, both of which directly affect the landmark-based baseline.

%
%

In this work, we use four main metrics: 
\begin{enumerate*}[label=(\roman*)]
  \item spread in goals ($S \in \goals$),
  \item accuracy~(Acc),
  \item precision~(Prec), and
  \item recognition time.
\end{enumerate*}
Spread is the average cardinality of the set of goal hypotheses returned by the recogniser as most likely correct ones.
Accuracy is the percentage of problems where the recogniser correctly identified the correct intended goal,~\idest~when the correct intended goal $\realgoal$ is included in the set of returned hypotheses. 
Since a recogniser that always returns the entire candidate goal set achieves perfect accuracy, we also report the average precision per problem,~\idest~the average fraction of returned hypotheses that are correct, by dividing the number of correctly recognised goals ($1$ if the model was correct, $0$ otherwise) by the spread in $\goals$. 
Finally, we report the recognition time in seconds as a measure of efficiency for the different models and the baseline.

For the LLM-based recognisers, we also report the token usage and cost in USD (for the commercial models) as additional metrics of interest. 
We report not only the total number of tokens (TT) used, but also the breakdown between input (PT) and output (CT) tokens. 
The idea behind these metrics is understanding which model is more efficient in terms of token usage versus recognition performance. 
We provide an overview of these metrics in Section~\ref{sec:results}, and the complete information on token usage and cost in Table~\ref{tab:tokens_all} in Appendix~\ref{app:tokens}.

To augment the quantitative analysis, we perform a qualitative analysis over the reasoning reported by the different models to better understand their behavior.
In it, we recognise how LLMs approach the problem, identify potential patterns in the reasoning that could explain the differences in performance between each model, and provide some common failure points in all models.
Due to space constraints, all examples for the qualitative analysisappear as part of Appendix~\ref{app:reasoning}.
Finally, we provide some potential points towards a productive agenda to improve LLMs further in goal recognition tasks.



\section{Results}
\label{sec:results}
%
%

\begin{table*}[!tp]
\normalsize
\centering
\renewcommand{\arraystretch}{1.3}
\setlength{\tabcolsep}{3pt}
\caption{
    Results for Landmark-Based, \qwen, \gptoss, \gptfour, and \gptfive goal recognition on all problems.
    ALL DOMAINS shows the average across all problems in the four domains.
}
\label{tab:gr_comparison_all}
\resizebox{\textwidth}{!}{
\begin{tabular}{cccccccccccccccccccccccccc}
\toprule
& & & & & &
\multicolumn{4}{c}{\textbf{Landmark-Based}} & 
\multicolumn{4}{c}{\textbf{Qwen3.5}} & 
\multicolumn{4}{c}{\textbf{GPT-OSS}} & 
\multicolumn{4}{c}{\textbf{GPT-4o}} & 
\multicolumn{4}{c}{\textbf{GPT-5.4}} \\
\cmidrule(lr){7-10}\cmidrule(lr){11-14}\cmidrule(lr){15-18}\cmidrule(lr){19-22}\cmidrule(lr){23-26}
\textbf{Domain} & 
$|\landmarks|$ & 
$\observations~\%$ & 
$|\problems|$ & 
$|\goals|$ & 
$|\observations|$ & 
$S {\in} \goals$ & \textbf{Acc} & \textbf{Prec} & \textbf{Time} & 
$S {\in} \goals$ & \textbf{Acc} & \textbf{Prec} & \textbf{Time} & 
$S {\in} \goals$ & \textbf{Acc} & \textbf{Prec} & \textbf{Time} & 
$S {\in} \goals$ & \textbf{Acc} & \textbf{Prec} & \textbf{Time} & 
$S {\in} \goals$ & \textbf{Acc} & \textbf{Prec} & \textbf{Time} \\
\cmidrule{1-1}\cmidrule{2-6}\cmidrule(lr){7-10}\cmidrule(lr){11-14}\cmidrule(lr){15-18}\cmidrule(lr){19-22}\cmidrule(lr){23-26}
\multirow{5}{*}{\rotatebox{90}{\makecell{BLOCKS\\WORLD}}} & \multirow{5}{*}{\rotatebox{90}{16.6}} & 10 & 246 & 20.3 & 1.8 & 1.3 & 44.31 & \textbf{38.04} & \textbf{2.9} & 0.0 & 0.41 & 0.41 & 517.6 & 6.6 & \textbf{63.10} & 20.23 & 201.4 & 5.3 & 43.50 & 14.05 & 12.1 & 6.4 & 53.25 & 11.63 & 16.7 \\
 &  & 30 & 246 & 20.3 & 4.9 & 1.2 & 56.10 & \textbf{51.02} & \textbf{2.8} & 0.0 & 0.41 & 0.41 & 536.2 & 4.6 & \textbf{63.41} & 32.51 & 137.5 & 2.6 & 30.49 & 13.66 & 12.1 & 2.5 & 36.59 & 18.80 & 17.0 \\
 &  & 50 & 246 & 20.3 & 7.7 & 1.1 & 63.41 & \textbf{58.88} & \textbf{2.2} & 0.0 & 1.63 & 1.42 & 675.7 & 3.5 & \textbf{69.51} & 43.71 & 126.1 & 2.2 & 33.33 & 19.50 & 10.9 & 2.0 & 45.93 & 27.28 & 17.6 \\
 &  & 70 & 246 & 20.3 & 11.1 & 1.2 & \textbf{83.74} & \textbf{78.32} & \textbf{1.9} & 0.0 & 2.03 & 2.03 & 538.7 & 2.5 & 80.49 & 63.21 & 135.8 & 2.6 & 34.15 & 18.15 & 10.8 & 1.6 & 54.07 & 39.72 & 18.0 \\
 &  & 100 & 92 & 20.3 & 14.5 & 1.4 & \textbf{100.00} & 85.69 & \textbf{1.7} & 0.3 & 22.83 & 21.74 & 671.8 & 1.4 & 98.91 & \textbf{94.68} & 121.2 & 2.0 & 41.30 & 26.18 & 10.8 & 1.5 & 77.17 & 59.87 & 16.0 \\
\cmidrule{1-1}\cmidrule{2-6}\cmidrule(lr){7-10}\cmidrule(lr){11-14}\cmidrule(lr){15-18}\cmidrule(lr){19-22}\cmidrule(lr){23-26}
\multirow{5}{*}{\rotatebox{90}{CAMPUS}} & \multirow{5}{*}{\rotatebox{90}{9.7}} & 10 & 15 & 2.0 & 1.0 & 1.1 & 86.67 & \textbf{80.00} & \textbf{2.8} & 1.0 & 80.00 & \textbf{80.00} & 204.5 & 1.4 & 93.33 & 73.33 & 27.1 & 1.9 & \textbf{100.00} & 53.33 & 5.7 & 1.5 & 93.33 & 70.00 & 6.6 \\
 &  & 30 & 15 & 2.0 & 2.0 & 1.1 & 86.67 & 80.00 & \textbf{2.9} & 0.1 & 0.00 & 0.00 & 407.5 & 1.1 & 86.67 & \textbf{86.67} & 32.1 & 1.7 & \textbf{100.00} & 66.67 & 4.9 & 1.5 & 93.33 & 70.00 & 8.4 \\
 &  & 50 & 15 & 2.0 & 3.0 & 1.1 & 93.33 & \textbf{86.67} & \textbf{2.8} & 0.2 & 6.67 & 6.67 & 475.4 & 1.1 & 86.67 & 80.00 & 29.0 & 1.4 & \textbf{100.00} & 80.00 & 7.2 & 1.3 & \textbf{100.00} & 83.33 & 7.9 \\
 &  & 70 & 15 & 2.0 & 4.5 & 1.0 & \textbf{100.00} & \textbf{100.00} & \textbf{2.9} & 1.0 & \textbf{100.00} & \textbf{100.00} & 202.8 & 1.0 & 86.67 & 86.67 & 28.8 & 1.3 & \textbf{100.00} & 86.67 & 5.7 & 1.1 & \textbf{100.00} & 96.67 & 9.7 \\
 &  & 100 & 15 & 2.0 & 5.4 & 1.0 & \textbf{100.00} & \textbf{100.00} & \textbf{2.8} & 0.0 & 0.00 & 0.00 & 466.4 & 1.1 & 86.67 & 83.33 & 25.9 & 1.1 & \textbf{100.00} & 96.67 & 6.1 & 1.0 & \textbf{100.00} & \textbf{100.00} & 9.0 \\
\cmidrule{1-1}\cmidrule{2-6}\cmidrule(lr){7-10}\cmidrule(lr){11-14}\cmidrule(lr){15-18}\cmidrule(lr){19-22}\cmidrule(lr){23-26}
\multirow{5}{*}{\rotatebox{90}{DRIVERLOG}} & \multirow{5}{*}{\rotatebox{90}{10.8}} & 10 & 84 & 7.1 & 2.6 & 1.3 & 45.24 & \textbf{40.58} & \textbf{4.4} & 0.0 & 0.00 & 0.00 & 538.8 & 2.8 & \textbf{65.48} & 36.65 & 57.6 & 2.7 & 45.24 & 16.92 & 9.3 & 2.3 & 44.05 & 20.26 & 15.0 \\
 &  & 30 & 84 & 7.1 & 7.0 & 1.2 & 59.52 & \textbf{51.98} & \textbf{4.4} & 0.0 & 0.00 & 0.00 & 514.4 & 2.0 & \textbf{70.24} & 48.38 & 79.9 & 1.8 & 30.95 & 17.75 & 9.4 & 1.3 & 41.67 & 31.79 & 15.9 \\
 &  & 50 & 84 & 7.1 & 11.2 & 1.3 & 77.38 & 67.36 & \textbf{4.4} & 0.4 & 22.62 & 20.83 & 588.3 & 1.4 & \textbf{86.90} & \textbf{75.58} & 74.2 & 1.5 & 44.05 & 28.27 & 9.4 & 1.1 & 53.57 & 47.42 & 16.5 \\
 &  & 70 & 84 & 7.1 & 15.6 & 1.2 & 92.86 & 84.13 & \textbf{4.3} & 1.0 & 80.95 & 80.36 & 346.4 & 1.2 & \textbf{95.24} & \textbf{90.67} & 75.2 & 1.7 & 47.62 & 30.99 & 9.6 & 1.2 & 73.81 & 63.29 & 16.3 \\
 &  & 100 & 28 & 7.1 & 21.7 & 1.2 & \textbf{100.00} & 90.48 & \textbf{4.1} & 0.0 & 0.00 & 0.00 & 498.8 & 1.0 & \textbf{100.00} & \textbf{100.00} & 50.0 & 1.3 & 46.43 & 35.71 & 8.5 & 1.2 & 96.43 & 83.93 & 14.0 \\
\cmidrule{1-1}\cmidrule{2-6}\cmidrule(lr){7-10}\cmidrule(lr){11-14}\cmidrule(lr){15-18}\cmidrule(lr){19-22}\cmidrule(lr){23-26}
\multirow{5}{*}{\rotatebox{90}{DWR}} & \multirow{5}{*}{\rotatebox{90}{32.0}} & 10 & 84 & 7.3 & 5.7 & 1.2 & 38.10 & \textbf{35.12} & \textbf{4.4} & 0.0 & 0.00 & 0.00 & 455.8 & 4.0 & \textbf{73.81} & 26.76 & 77.6 & 2.3 & 35.71 & 16.26 & 8.4 & 2.2 & 25.00 & 9.07 & 16.2 \\
 &  & 30 & 84 & 7.3 & 16.0 & 1.1 & 64.29 & \textbf{61.31} & \textbf{4.4} & 0.7 & 34.52 & 30.06 & 645.2 & 2.5 & \textbf{78.57} & 50.10 & 105.5 & 1.5 & 25.00 & 17.26 & 9.3 & 1.4 & 23.81 & 15.16 & 16.8 \\
 &  & 50 & 84 & 7.3 & 26.2 & 1.1 & 72.62 & \textbf{70.24} & \textbf{4.5} & 0.1 & 5.95 & 5.95 & 527.9 & 2.0 & \textbf{85.71} & 59.47 & 110.9 & 1.8 & 26.19 & 15.97 & 8.3 & 1.1 & 26.19 & 21.73 & 17.0 \\
 &  & 70 & 84 & 7.3 & 36.9 & 1.0 & \textbf{90.48} & \textbf{88.10} & \textbf{3.8} & 0.1 & 3.57 & 3.57 & 567.7 & 1.5 & \textbf{90.48} & 78.08 & 131.6 & 1.5 & 26.19 & 19.35 & 8.5 & 1.1 & 30.95 & 26.79 & 17.1 \\
 &  & 100 & 28 & 7.3 & 51.9 & 1.0 & \textbf{92.86} & \textbf{92.86} & \textbf{3.6} & 0.8 & 67.86 & 67.86 & 524.8 & 1.0 & \textbf{92.86} & \textbf{92.86} & 105.6 & 1.9 & 25.00 & 13.07 & 7.8 & 1.0 & 53.57 & 51.79 & 14.7 \\
\cmidrule{1-1}\cmidrule{2-6}\cmidrule(lr){7-10}\cmidrule(lr){11-14}\cmidrule(lr){15-18}\cmidrule(lr){19-22}\cmidrule(lr){23-26}
\multirow{5}{*}{\rotatebox{90}{\makecell{ALL\\DOMAINS}}} & \multirow{5}{*}{\rotatebox{90}{17.3}} & 10 & 429 & 14.5 & 2.7 & 1.3 & 44.76 & \textbf{39.43} & \textbf{3.5} & 0.1 & 3.03 & 3.03 & 498.7 & 4.9 & \textbf{67.30} & 27.59 & 133.6 & 4.1 & 44.29 & 16.42 & 10.6 & 4.6 & 47.32 & 14.86 & 15.9 \\
 &  & 30 & 429 & 14.5 & 7.4 & 1.2 & 59.44 & \textbf{54.23} & \textbf{3.5} & 0.1 & 6.99 & 6.12 & 548.8 & 3.5 & \textbf{68.53} & 40.96 & 116.3 & 2.2 & 31.93 & 17.02 & 10.7 & 2.0 & 37.06 & 22.42 & 16.5 \\
 &  & 50 & 429 & 14.5 & 11.8 & 1.1 & 69.00 & \textbf{63.73} & \textbf{3.1} & 0.1 & 6.76 & 6.29 & 622.7 & 2.7 & \textbf{76.69} & 54.30 & 109.6 & 2.0 & 36.36 & 22.64 & 10.0 & 1.7 & 45.45 & 32.09 & 16.9 \\
 &  & 70 & 429 & 14.5 & 16.8 & 1.1 & \textbf{87.41} & \textbf{82.13} & \textbf{2.8} & 0.3 & 21.21 & 21.10 & 495.0 & 2.0 & 85.55 & 72.32 & 119.4 & 2.2 & 37.53 & 23.29 & 9.9 & 1.4 & 55.01 & 43.79 & 17.2 \\
 &  & 100 & 163 & 14.1 & 21.3 & 1.2 & \textbf{98.77} & 89.06 & \textbf{2.5} & 0.3 & 24.54 & 23.93 & 597.9 & 1.3 & 96.93 & \textbf{94.23} & 97.5 & 1.8 & 44.79 & 32.05 & 9.4 & 1.3 & 78.53 & 66.31 & 14.8 \\
\bottomrule
\end{tabular}
}
\end{table*}

Table~\ref{tab:gr_comparison_all} summarises accuracy, precision, spread, and time in seconds for all five recognisers across the four evaluation domains.
The central result is that the recognisers divide into three qualitatively distinct groups.
\gptoss leads all LLMs at every observation percentage, achieving $67.3\%$ accuracy at $10\%$ observations across all domains, surpassing landmark-based in accuracy ($44.8\%$), and scaling to $96.9\%$ at full observations, nearly matching landmark-based ($98.8\%$).
\gptfive shows partial scaling, rising from $47.3\%$ to $78.5\%$ in accuracy, but falls short of \gptoss throughout.
\gptfour, by contrast, remains effectively flat: $44.3\%$ in accuracy at $10\%$ observations and $44.8\%$ at $100\%$, a range of less than one percentage point despite a tenfold increase in evidence.
\qwen fails almost entirely on all but the simplest domain, reaching only $24.5\%$ at full observations across all domains.

The accuracy gap between \gptfour and \gptoss is clearest on \emph{Blocks World}, where landmark-based and \gptfour are tied at $10\%$ observations ($44.3\%$ and $43.5\%$ respectively), but diverge completely thereafter: landmark-based reaches $100\%$ at full observations, \gptoss reaches $96.9\%$, and \gptfour never exceeds $43.5\%$.
On \emph{DWR}, the hardest domain, \gptoss reaches $92.9\%$ at full observations, matching landmark-based exactly, while \gptfour remains near $25\%$ regardless of evidence.
\emph{Campus}, the simplest domain, is the exception: \gptfour achieves $100\%$ accuracy at all observability degrees, suggesting that world knowledge alone suffices when the goal space is small ($|\goals|=2$) and the vocabulary familiar.
\qwen fails even on campus at $30\%$ and $100\%$ observations, indicating that local inference at this scale is insufficient for reliable output formatting, not merely for reasoning.

\begin{figure}[!tp]
\centering
\includegraphics[width=1.0\linewidth]{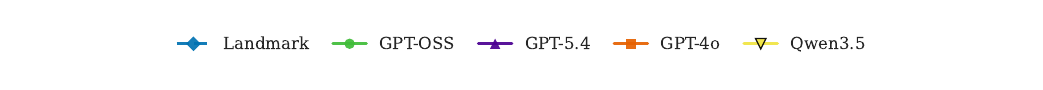}
\includegraphics[width=0.31\linewidth]{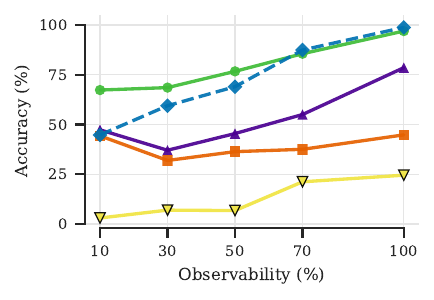}
\hfill
\includegraphics[width=0.31\linewidth]{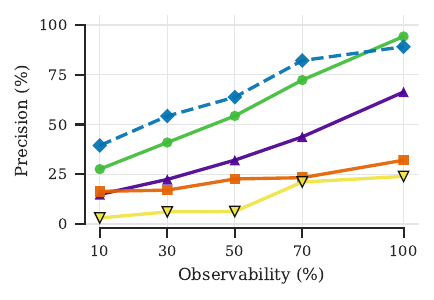}
\hfill
\includegraphics[width=0.31\linewidth]{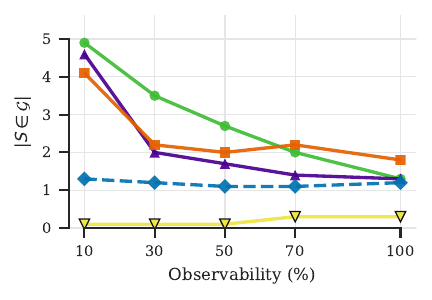}
\caption{%
    Average acc, prec, and spread for different observation percentages for ALL DOMAINS.
}
\label{fig:line-plots}
\end{figure}

Accuracy alone is an incomplete picture because a recogniser returning all candidate goals achieves perfect recall at the cost of zero discriminative power.
Spread and precision track this tradeoff directly.
We use the spread to compute the precision on each problem, dividing the number of correctly identified goals (which can be at most one) by the number of returned hypotheses.
At $10\%$ observations, \gptoss and \gptfour start with similarly high spread ($4.9$ and $4.1$ respectively, versus $1.3$ for landmark-based), reflecting that both LLMs hedge broadly when evidence is sparse.
Even with a higher spread, \gptoss achieves $27.6\%$ precision at $10\%$ observations due to its higher accuracy, compared to $16.4\%$ for \gptfour and $39.4\%$ for landmark-based.
As observations accumulate, \gptoss resolves its uncertainty: spread falls to $1.3$ at full observations, approaching landmark-based ($1.2$), and precision rises to $94.2\%$, the best overall on full observations.
\gptfour's precision increase for higher observability is solely due to its spread decreasing, as its accuracy remains flat.
Its spread decreases only to $1.8$ and accuracy remains flat at full observations, so precision reaches just $32.1\%$, compared to $89.1\%$ for landmark-based and $94.2\%$ for \gptoss.
The precision gap between \gptoss and \gptfour at full observations is thus as striking as the accuracy gap: both can retrieve the correct goal, but only \gptoss commits to it.
On the other hand, \gptfive shows a more balanced improvement in both accuracy and spread, leading to a steady increase in precision from $14.9\%$ at $10\%$ observations to $66.3\%$ at $100\%$.
Similarly to its accuracy results, \qwen's precision is also low across all observation percentages, reaching only $23.9\%$ at full observations.
We also see that \qwen's spread's average is mostly below $1$, due to the fact that it often fails completely and returns no candidate goals at all.
Figure~\ref{fig:line-plots} shows these trends more clearly.
We discuss the reasoning inconsistencies that lead to this kind of anomaly in the results in Section~\ref{sec:discussion}.

In terms of recognition time, the landmark-based baseline is by far the most efficient, with its average time ranging from $1.7$ to $4.5$ seconds across all domains and observation percentages.
Among the LLM-based recognisers, \gptfour is the fastest one, with the average recognition time ranging from $4.9$ to $12.1$ seconds, followed by \gptfive. \qwen and \gptoss are the slowest, which is due to the fact that they are non-commercial models running on local hardware.

Regarding token usage, Table~\ref{tab:tokens_all_domains} shows that \gptfour is overall the most efficient model in terms of the usage of completion tokens, while \gptfive consumes less prompt tokens.
\gptfour dominates as the most token-efficient model, with a total average of around $3{,}000$ tokens per problem, compared to around $3{,}300$ for \gptfive.
\gptoss and \qwen are the least efficient, with $11{,}500$ total tokens used for \gptoss, and $18{,}700$ for \qwen.
In terms of cost in the commercial models, all of them kept the average cost per problem around $0.01$ and $0.02$ USD.
We report the complete information on the cost of commercial models in terms of the average number of input and output tokens, as well as the average cost per recognition, in Table~\ref{tab:tokens_all} in Appendix~\ref{app:tokens}.

\begin{table}[b!tp]
\normalsize
\centering
\renewcommand{\arraystretch}{1.3}
\setlength{\tabcolsep}{3pt}
\caption{
    Token usage and cost in USD (avg.\ per problem) for LLMs on all domains.
}
\label{tab:tokens_all_domains}
\resizebox{\textwidth}{!}{
\begin{tabular}{ccccc ccc ccc cccc cccc}
\toprule
& & & & &
\multicolumn{3}{c}{\textbf{Qwen3.5}} &
\multicolumn{3}{c}{\textbf{GPT-OSS}} &
\multicolumn{4}{c}{\textbf{GPT-4o}} &
\multicolumn{4}{c}{\textbf{GPT-5.4}} \\
\cmidrule(lr){6-8}\cmidrule(lr){9-11}\cmidrule(lr){12-15}\cmidrule(lr){16-19}
\textbf{Domain} & $\observations~\%$ & $|\problems|$ & $|\goals|$ & $|\observations|$ &
\textbf{PT} & \textbf{CT} & \textbf{TT} &
\textbf{PT} & \textbf{CT} & \textbf{TT} &
\textbf{PT} & \textbf{CT} & \textbf{TT} & \textbf{Cost} &
\textbf{PT} & \textbf{CT} & \textbf{TT} & \textbf{Cost} \\
\cmidrule{1-1}\cmidrule{2-5}\cmidrule(lr){6-8}\cmidrule(lr){9-11}\cmidrule(lr){12-15}\cmidrule(lr){16-19}
\multirow{5}{*}{\rotatebox{90}{\makecell{ALL\\DOMAINS}}} & 10 & 429 & 14.5 & 2.7 & 2135.9 & 13656.4 & 15792.3 & \textbf{2003.3} & 6824.3 & 8827.6 & 2031.0 & \textbf{956.0} & \textbf{2987.0} & 0.01 & 2030.0 & 1216.4 & 3246.4 & 0.02 \\
 & 30 & 429 & 14.5 & 7.4 & 2202.2 & 16258.8 & 18460.9 & 2155.0 & 10301.8 & 12456.7 & 2089.8 & \textbf{918.5} & \textbf{3008.3} & 0.01 & \textbf{2088.8} & 1220.2 & 3309.0 & 0.02 \\
 & 50 & 429 & 14.5 & 11.8 & 2283.6 & 18912.6 & 21196.2 & 2211.3 & 10626.4 & 12837.6 & 2146.1 & \textbf{903.5} & \textbf{3049.6} & 0.01 & \textbf{2145.1} & 1241.6 & 3386.6 & 0.02 \\
 & 70 & 429 & 14.5 & 16.8 & 2354.5 & 15529.1 & 17883.7 & 2273.6 & 11220.5 & 13494.0 & 2208.4 & \textbf{914.2} & \textbf{3122.6} & 0.01 & \textbf{2207.4} & 1232.5 & 3439.9 & 0.02 \\
 & 100 & 163 & 14.1 & 21.3 & 2395.4 & 18072.7 & 20468.0 & 2331.3 & 7836.0 & 10167.4 & 2266.3 & \textbf{882.9} & \textbf{3149.1} & 0.01 & \textbf{2265.3} & 1127.8 & 3393.0 & 0.02 \\
\bottomrule
\end{tabular}
}
\end{table}

\nl{Token usage is (potentially) useful but I'm not convinced by the cost element.}
\nsg{I think we can remove PC from the table. It doesn't add much to it, and since PT is just TT - CT, readers can easily compute it if they want.}
\nl{At some point there was a discussion about runtime. There's an obvious question about how this scales for the different approaches as the size of the (same) domain increases.}


\section{Discussion}
\label{sec:discussion}

Given the quantitative results, we now focus on analysing the reasoning process of LLMs when performing goal recognition, and provide some examples where we believe improvements can be made to allow LLMs to perform better.
We select examples from the \emph{Blocks-World} domain, as it is a commonly used domain in the goal recognition literature, and one that is adopted for the planning experiments in \citeauthor{CorreaPereiraSeipp2025a}'s work [\citeyear{CorreaPereiraSeipp2025a}]. We reference figures containing the reasoning excerpts in Appendix~\ref{app:reasoning} for optional review.

\vspace{-0.25cm}
\paragraph{\gptfour}
\gptfour's reasoning is consistently inconsistent with its own score assignments, a pattern that holds across all observability levels.
At low observability, the model tends to anchor on the few observed actions and check which hypotheses include the affected blocks, but does so overconfidently: in Figure~\ref{fig:gpt-4o-correct-reasoning-example-1}, it selects a single hypothesis despite multiple others being equally compatible with the observation, and its justification contains factual errors about the preconditions.
This overconfidence is compounded by \emph{confabulation}, where the model produces plausible-sounding justifications that directly contradict its own scores: in Figure~\ref{fig:gpt-4o-correct-reasoning-example-2} it claims no hypothesis mentions the relevant block, when several do; in Figure~\ref{fig:gpt-4o-wrong-reasoning-example-1} it states it favoured hypotheses involving block \texttt{P}, yet the correct hypothesis does not prominently include \texttt{P}; and in Figure~\ref{fig:gpt-4o-correct-but-wrong-reasoning-example-1} the reasoning identifies discriminating facts and then collapses to a uniform distribution without explanation.
At higher observability, the failure mode shifts but the underlying problem persists: in Figure~\ref{fig:gpt-4o-wrong-reasoning-example-4} the reasoning names block \texttt{P} as relevant but the scores ignore it; in Figure~\ref{fig:gpt-4o-wrong-reasoning-example-3} the model identifies a relation between blocks \texttt{A} and \texttt{J} that is not supported by the observations; and in Figure~\ref{fig:gpt-4o-wrong-reasoning-example-2} the conclusion abruptly introduces blocks \texttt{M} and \texttt{O}, neither present in the observations nor in the preceding reasoning, to justify assigning maximum score to an incorrect hypothesis.
Across all cases, evidence that is named in the reasoning is not enforced in the scores, and evidence that drives the scores is not grounded in the reasoning.

\vspace{-0.25cm}
\paragraph{\gptfive}
\gptfive exhibits more sophisticated planning reasoning than \gptfour, but this comes with its own failure modes.
Unlike \gptfour, it identifies precondition violations arising from partial observability and reasons around them rather than ignoring them: in Figure~\ref{fig:gpt-5-4-correct-reasoning-example-1}, it correctly infers that two of the three observed actions jointly establish \texttt{(ON P A)}, identifies the unique hypothesis containing that fact, and selects it correctly.
It also reasons about hypotheses as goals that the observed sequence may be partially progressing toward, which is a principled strategy under partial observability, as illustrated in Figure~\ref{fig:gpt-5-4-wrong-reasoning-example-2}, where it narrows to two candidates and selects the one most consistent with the observation as a partial plan.
However, this same tendency to commit early becomes a liability when evidence is insufficient or the problem is complex: under full observability it selects the wrong hypothesis in Figure~\ref{fig:gpt-5-4-wrong-reasoning-example-2} by misreading a completed sequence as an incomplete one, it derives an entirely incorrect $10$-block final stack in Figure~\ref{fig:gpt-5-4-wrong-reasoning-example-1} despite having all actions available, and in Figure~\ref{fig:gpt-5-4-wrong-reasoning-example-3} it overcommits to a single hypothesis at $30\%$ observability despite initially narrowing correctly to a set of four candidates that included the correct goal.
The net effect is higher precision when the reasoning is right, but lower accuracy when it is not.

\vspace{-0.25cm}
\paragraph{\gptoss}
\gptoss displays a recurring behaviour of applying a fact-matching strategy in most problems, trying to check the actual facts present in the state resulting from the execution of the observed actions. 
In Figure~\ref{fig:gpt-oss-correct-reasoning-example-1}, the model correctly identifies the correct hypothesis by reasoning about what are the facts present in the state following the applying of the single observed action, and assigns to each hypothesis a score proportional to the number of matched facts.
By doing so, it is able to obtain more consistent reasoning, but this behaviour still leads it to incorrect inference many times, like in the example from Figure~\ref{fig:gpt-oss-wrong-reasoning-example-1}.
In this example, the model does its best with the provided information, and correctly identifies the hypothesis that best matches the observed actions, but since there is lacking information due to partial observability, the model interprets an intermediate state as a final one.
Since in the reasoning it correctly identifies the missing information by stating that it is ignoring the preconditions being violated, it could in theory try to better distribute the score to include a larger number of matching hypotheses.
Different from the previous example, where the model explicitly informed about ignoring the violated preconditions, in another problem the same model stated that since the observations contained precondition violations, it could not infer the correct intended goal, and hence assigned a uniform score to all hypotheses (Figure~\ref{fig:gpt-oss-wrong-reasoning-example-2}).
This showcases an inconsistency in the model's behaviour, which is expected to some degree given the nature of LLMs.
Additionally, it is interesting to see how it deals with the same issue differently, considering that the prompt does not specify that the observation sequence can be a non-contiguous partial subset of the full observation sequence.

\vspace{-0.25cm}
\paragraph{\qwen}
Out of all four evaluated models, \qwen was the worst when it comes to reasoning inconsistencies and hallucinations.
For many problems across all domains and observability degrees, the model did not provide any response at all, returning a completely empty output, even while consuming tokens.
In other cases, it correctly identifies a precondition violation under partial observability, but in turn simply states the sequence of actions is impossible and refuses to provide any further response, as in the example from Figure~\ref{fig:qwen3-5-wrong-reasoning-example-1}.
In a specific problem, \qwen provides unsound reasoning, showcasing explicit confusion and even confusing block names.
It also fails to deliver a response in the required format and select a hypothesis that was not even in the hypothesis list (Figure~\ref{fig:qwen3-5-wrong-reasoning-example-2}).
In another problem, with $30\%$ observability, the model fails completely and generates a nonsensical response, with its only response being ``You put x on table.'' as shown in Figure~\ref{fig:qwen3-5-wrong-reasoning-example-3}.
Another example of this kind of behaviour from \qwen is shown in Figure~\ref{fig:qwen3-5-wrong-reasoning-example-4}, where the model provides a completely incoherent response comprised only of ``[:answer A]'' under full observability.
For both cases, the model used around $12{,}000$ completion tokens, meaning there was reasoning effort but still no sound response.
Still, there are cases where \qwen shows reasoning capabilities and even correctly narrows down to a single and correct hypothesis under sensible reasoning, as in the example from Figure~\ref{fig:qwen3-5-correct-reasoning-example-1}.
In this example, even under $50\%$ observability, the model correctly identifies the resulting state from the observed actions and narrows down to two possibly correct hypotheses.
After that, it identifies an inconsistency between one of the selected hypotheses and the initial state, unchanged by the observations, and hence correctly selects the other hypothesis as the most likely correct one.

\vspace{-0.25cm}
\paragraph{Common Patterns}
Despite their differences in capability and failure mode, the models examined here share a consistent set of weaknesses that together point toward a productive research agenda.
All models struggle to translate stated reasoning into consistent score assignments, all exhibit overconfidence when observations are sparse, and none reliably improve their predictions as additional evidence accumulates, which is the defining discriminative demand of goal recognition.
All models also display position bias consistent with `lost-in-the-middle'~\citep{liu2024lost}: attention degrades for hypotheses placed in the middle of the candidate list, with \gptfour exhibiting a pronounced tendency to favour hypotheses at the first position.
Yet the broader picture is not discouraging.
The models collectively demonstrate that world knowledge of planning domains transfers meaningfully to zero-shot goal recognition: even without fine-tuning, all models perform non-trivially at low observability, and the progression from \gptfour to \gptfive shows that stronger general reasoning capabilities do address specific failure modes, notably precondition awareness and partial-plan reasoning.
This suggests that the bottleneck is not knowledge but the ability to integrate evidence incrementally.
Several targeted interventions could plausibly close the remaining gap.
Restructuring the output format to elicit reasoning before scores would allow deliberate comparison of candidate goals to genuinely constrain the prediction.
Training models to maintain an explicit running state representation updated action by action would directly address the state-tracking failures observed at higher observability.
Improving calibration so that models express uncertainty proportional to the available evidence would reduce the precision-accuracy trade-off that characterises \gptfive's behaviour.
A systematic prompt sensitivity analysis varying hypothesis ordering and position would further quantify the extent of position bias and inform more robust evaluation protocols~\citep{valmeekam2023planbench}.
More broadly, goal recognition provides an unusually clean benchmark for the foundational planning knowledge of LLMs: the ground truth is unambiguous, the evidence is structured and incremental, and the failure modes are precisely measurable.
The plateau at low observability reveals the floor set by world-knowledge priors; the divergence in scaling behaviour across models reveals how far that foundational knowledge alone can carry recognition and where evidence integration becomes the limiting factor.

\section{Related Work}
\label{sec:related}
The relationship between LLMs and automated planning is one of the most actively studied questions in recent AI research.
Early systematic evaluation using the PlanBench benchmark shows that LLMs fail to reliably solve classical planning problems; on standard IPC domains zero-shot performance is near chance, and it collapses entirely when domain and object names are obfuscated to remove semantic cues~\cite{ValmeekamOlmoSreedharanEtAl2022,ValmeekamMarquezSreedharanEtAl2023}.
\citet{KambhampatiValmeekamGuanEtAl2024} argue against using LLMs as standalone planners and propose the LLM-Modulo framework, which embeds LLMs within a loop controlled by external formal verifiers.
A complementary finding is that LLMs cannot reliably verify their own planning outputs~\cite{StechlyValmeekamKambhampati2025}.
More recent benchmarking finds that frontier LLMs and reasoning models have closed much of the gap with classical planners on standard IPC domains~\cite{valmeekam2025a,CorreaPereiraSeipp2025a}; obfuscated variants continue to expose substantial performance drops, confirming that this progress rests on world-knowledge exploitation rather than symbolic search.

\citet{ChiariGereviniPercassiEtAl2023} introduce GRNet, which recasts goal recognition as a deep learning classification task: a neural network trained on plan traces from a given domain learns to predict the goal from a partial observation sequence.
\citet{SerinaChiariGereviniEtAl2025} extend this line to online settings, where the recogniser updates its prediction incrementally as new observations arrive.
These approaches achieve high accuracy and fast inference but require domain-specific training data, making them unsuitable for zero-shot evaluation across novel domains.
A distinct line of work addresses goal recognition under noisy or incomplete observations using neuro-symbolic methods~\cite{PereiraFuggittiMeneguzziEtAl2023,TeschAmadoMeneguzzi2024}; that setting differs from ours, in which observations are assumed complete but partial.
Our contribution is complementary to both lines: we ask whether world knowledge encoded in pre-trained LLMs suffices for zero-shot goal recognition, without any domain-specific training.


\section{Conclusion}
\label{sec:conclusion}
%
%

This paper provides the first systematic evaluation of LLMs in the task of goal recognition over PDDL benchmarks. 
Given the substantial challenge~\cite{ValmeekamMarquezSreedharanEtAl2023}, and recently perceived progress~\cite{CorreaPereiraSeipp2025a}, these benchmarks provide a closely related, but qualitatively different challenge to language models, and how they employ world knowledge.
Our empirical experiments show that even models that approximate the performance of symbolic algorithms for planning (GPT-5.4) collapse when performing goal recognition in the same domain. 
These failures happen even without any obfuscation with three qualitatively distinct profiles. 
\gptoss leads all models at every observation percentage, scaling from $67.3\%$ to $96.9\%$ accuracy and approaching \textsc{lm} ($98.8\%$) at full observations. 
\gptfive shows partial scaling ($47.3\%$ to $78.5\%$). 
\gptfour remains flat at $44\%$ regardless of evidence. 
\qwen fails almost entirely due to output-format failures at this model scale.

The key discriminator between profiles is not domain knowledge but the ability to integrate additional evidence: \gptfour matches \textsc{lm} at $10\%$ observations, where world-knowledge priors suffice, then diverges completely as evidence accumulates. 
This is the predicted failure mode from Section~\ref{sec:method:complexity}: world-knowledge retrieval masquerading as goal recognition. 
Here \gptfour indicates that more domain knowledge acts as a disadvantage if the models do not perform reasoning.
Our prompt places likelihood scores before the free-text reasoning block as a deliberate methodological control.
It requires each model to commit to a distribution over goals using only its domain knowledge and the observation evidence, preventing the reasoning trace from post-hoc steering the scores toward a rationalised answer.
\gptfour's flat accuracy curve therefore reflects a genuine failure to integrate evidence rather than an artefact of the model being steered toward prior-anchored scores by its own reasoning. 
Whether reversing the order, reasoning first and scoring after, would improve evidence integration for models like \gptfour is an open question and a natural next step.

Goal recognition provides a principled benchmark for LLM reasoning: the accuracy floor at sparse observations measures the domain familiarity encoded in training, while the scaling slope over increasing observations reveals whether a model integrates evidence or merely retrieves priors. The contrast between \gptfour (flat slope, high floor) and \gptoss (rising slope) makes this separation empirically clean.
Several directions remain open: (i) a mystery-domain ablation to test whether \gptoss's scaling survives name obfuscation~\cite{ValmeekamOlmoSreedharanEtAl2022}; (ii) chain-of-thought prompting with reasoning placed before scores, to test whether evidence integration improves; and (iii) evaluation under noisy observations, where some observed actions are incorrect, a setting in which LLMs' approximate reasoning may provide an advantage over classical methods that assume noiseless input.

\begin{ack}
\frm[inline]{Acknowledgements. Remove this section from the anonymised
submission version.}
\end{ack}


\bibliographystyle{plainnat}
\bibliography{references}

\clearpage
\appendix


\section{Prompt Template}
\label{app:prompt}

\begin{figure}[H]
\begin{lstlisting}[style=prompt]
You are an expert at goal recognition in AI planning using planning domain descriptions provided in PDDL format.
You will be provided with the following: 

- Planning domain description in PDDL. This contains the definition of the possible types of objects and the description of the actions that can be performed in the environment. Each action is comprised by the following:
    - A name
    - A list of parameters (with their types)
    - A list of precondition represented as a logical AND of facts, meaning facts that must hold in the state for the action to be applicable
    - A list of effects represented as a logical AND of facts, meaning facts that will be added to the state or removed from the state if they are preceded by the reserved word "not" after applying the action to the state
- Problem template, which is the PDDL description of the corresponding planning problem with the initial state and a placeholder for the goal state, comprising the following:
    - A list of objects in the problem with their types
    - An initial state represented as a simple list of facts separated by new line characters that hold in the initial state of the problem
    - A placeholder for the goal state which would be represented as a logical AND of facts, meaning facts that must hold in the final and goal state of the problem
- A list of goal hypotheses where one of them is the correct goal. In this list each goal is a list of facts. Each fact in a goal is separated by comma "," characters. Each goal hypothesis in the list is separated by new line characters.
- A list of observed actions that were performed by the agent. This is a simple list of executed actions separated by new line characters.

Your task is to analyze the problem provided to you and determine which goals among the goal hypotheses are the most likely to be the correct ones, given the observed actions of an agent.
For each goal hypothesis, assign a likelihood score (real number between 0 and 1) that represents how likely it is that the observed agent is pursuing that goal, based on the actions observed. This score MUST be normalized, so the sum of the likelihood scores of all goal hypotheses must be equal to 1. After assigning a likelihood score to each goal hypothesis, determine the set of goals which are tied with the highest likelihood score and provide an explanation of the reasoning process that led you to assign those likelihood scores and determine that set of most likely goals.
The output should be the list of goals along with their likelihood scores, the set of goals which are tied with the highest likelihood score, and after that an explanation of the reasoning process that led to the final answer. The format of the output should be exactly as the one specified below the "## Output Format" section of this prompt, without any additional text or formatting.

## Domain
{problem.domain}

## Problem Template
{problem.template}

## Goal Hypotheses
{goals_text}

## Observed Actions (in order)
{obs_text}

## Output Format (follow this format exactly, without any additional text or formatting):

Hyps:
Hyp: <goal hypothesis> | Score: <likelihood score between 0 and 1>

Most Likely Goals:
- <goal hypothesis 1>
- <goal hypothesis 2>
...

Reasoning:
<explanation of the reasoning process that led to the final answer>
\end{lstlisting}
\caption{Prompt template used for zero-shot goal recognition. Placeholders \texttt{\{problem.domain\}}, \texttt{\{problem.template\}}, \texttt{\{goals\_text\}}, and \texttt{\{obs\_text\}} are filled at runtime with the PDDL domain, problem template, goal hypotheses, and observed action sequence respectively.}
\label{fig:prompt-template}
\end{figure}

\section{Token Usage and Cost}
\label{app:tokens}

\clearpage
\begin{table*}[!tp]
\normalsize
\centering
\renewcommand{\arraystretch}{1.3}
\setlength{\tabcolsep}{3pt}
\caption{\footnotesize Token usage and cost in USD (avg.\ per problem) for LLMs on all problems.}
\label{tab:tokens_all}
\resizebox{\textwidth}{!}{
\begin{tabular}{ccccc ccc ccc cccc cccc}
\toprule
& & & & &
\multicolumn{3}{c}{\textbf{Qwen3.5}} &
\multicolumn{3}{c}{\textbf{GPT-OSS}} &
\multicolumn{4}{c}{\textbf{GPT-4o}} &
\multicolumn{4}{c}{\textbf{GPT-5.4}} \\
\cmidrule(lr){6-8}\cmidrule(lr){9-11}\cmidrule(lr){12-15}\cmidrule(lr){16-19}
\textbf{Domain} & $\observations~\%$ & $|\problems|$ & $|\goals|$ & $|\observations|$ &
\textbf{PT} & \textbf{CT} & \textbf{TT} &
\textbf{PT} & \textbf{CT} & \textbf{TT} &
\textbf{PT} & \textbf{CT} & \textbf{TT} & \textbf{Cost} &
\textbf{PT} & \textbf{CT} & \textbf{TT} & \textbf{Cost} \\
\cmidrule{1-1}\cmidrule{2-5}\cmidrule(lr){6-8}\cmidrule(lr){9-11}\cmidrule(lr){12-15}\cmidrule(lr){16-19}
\multirow{5}{*}{\rotatebox{90}{\makecell{BLOCKS\\WORLD}}} & 10 & 246 & 20.3 & 1.8 & 1862.6 & 14603.6 & 16466.2 & \textbf{1569.6} & 7441.7 & 9011.3 & 1766.5 & \textbf{1107.5} & \textbf{2874.0} & 0.02 & 1765.5 & 1363.9 & 3129.4 & 0.02 \\
 & 30 & 246 & 20.3 & 4.9 & 1903.4 & 15119.8 & 17023.2 & 1865.0 & 11664.3 & 13529.3 & 1796.0 & \textbf{1060.3} & \textbf{2856.3} & 0.02 & \textbf{1795.0} & 1329.0 & 3124.0 & 0.02 \\
 & 50 & 246 & 20.3 & 7.7 & 1933.6 & 20369.4 & 22303.0 & 1890.7 & 12266.6 & 14157.2 & 1821.7 & \textbf{1042.2} & \textbf{2863.9} & 0.01 & \textbf{1820.7} & 1357.7 & 3178.4 & 0.02 \\
 & 70 & 246 & 20.3 & 11.1 & 1972.0 & 14926.9 & 16898.9 & 1923.3 & 12590.8 & 14514.1 & 1854.3 & \textbf{1065.7} & \textbf{2920.0} & 0.02 & \textbf{1853.3} & 1335.5 & 3188.8 & 0.02 \\
 & 100 & 92 & 20.3 & 14.5 & 1995.2 & 18649.6 & 20644.8 & 1940.1 & 8975.4 & 10915.5 & 1871.1 & \textbf{1035.4} & \textbf{2906.5} & 0.02 & \textbf{1870.1} & 1244.7 & 3114.8 & 0.02 \\
\cmidrule{1-1}\cmidrule{2-5}\cmidrule(lr){6-8}\cmidrule(lr){9-11}\cmidrule(lr){12-15}\cmidrule(lr){16-19}
\multirow{5}{*}{\rotatebox{90}{CAMPUS}} & 10 & 15 & 2.0 & 1.0 & 2606.9 & 10672.1 & 13279.1 & 2492.5 & 2474.7 & 4967.2 & 2432.5 & \textbf{350.1} & \textbf{2782.7} & 0.01 & \textbf{2431.5} & 469.2 & 2900.7 & 0.01 \\
 & 30 & 15 & 2.0 & 2.0 & 2621.7 & 10285.3 & 12907.0 & 2505.7 & 2994.6 & 5500.3 & 2445.7 & \textbf{373.8} & \textbf{2819.5} & 0.01 & \textbf{2444.7} & 538.3 & 2983.1 & 0.01 \\
 & 50 & 15 & 2.0 & 3.0 & 2633.9 & 12413.7 & 15047.5 & 2516.9 & 2644.5 & 5161.4 & 2456.9 & \textbf{376.4} & \textbf{2833.3} & 0.01 & \textbf{2455.9} & 579.2 & 3035.1 & 0.01 \\
 & 70 & 15 & 2.0 & 4.5 & 2655.6 & 10565.6 & 13221.2 & 2537.1 & 2651.4 & 5188.5 & 2477.1 & \textbf{367.5} & \textbf{2844.7} & 0.01 & \textbf{2476.1} & 606.4 & 3082.5 & 0.02 \\
 & 100 & 15 & 2.0 & 5.4 & 2667.3 & 11792.2 & 14459.5 & 2548.9 & 2389.1 & 4937.9 & 2488.9 & \textbf{432.3} & \textbf{2921.1} & 0.01 & \textbf{2487.9} & 607.4 & 3095.3 & 0.02 \\
\cmidrule{1-1}\cmidrule{2-5}\cmidrule(lr){6-8}\cmidrule(lr){9-11}\cmidrule(lr){12-15}\cmidrule(lr){16-19}
\multirow{5}{*}{\rotatebox{90}{DRIVERLOG}} & 10 & 84 & 7.1 & 2.6 & 2768.2 & 13239.7 & 16007.9 & 2726.0 & 5542.3 & 8268.3 & 2666.0 & \textbf{850.3} & \textbf{3516.3} & 0.02 & \textbf{2665.0} & 1118.9 & 3783.9 & 0.02 \\
 & 30 & 84 & 7.1 & 7.0 & 2825.6 & 13143.2 & 15968.8 & 2795.2 & 7671.5 & 10466.8 & 2735.2 & \textbf{824.5} & \textbf{3559.7} & 0.02 & \textbf{2734.2} & 1161.8 & 3896.0 & 0.02 \\
 & 50 & 84 & 7.1 & 11.2 & 2989.7 & 17850.1 & 20839.8 & 2862.6 & 7111.2 & 9973.8 & 2802.6 & \textbf{803.1} & \textbf{3605.8} & 0.02 & \textbf{2801.6} & 1164.8 & 3966.4 & 0.02 \\
 & 70 & 84 & 7.1 & 15.6 & 3068.1 & 18628.5 & 21696.7 & 2933.7 & 7235.7 & 10169.4 & 2873.7 & \textbf{813.8} & \textbf{3687.5} & 0.02 & \textbf{2872.7} & 1166.8 & 4039.5 & 0.02 \\
 & 100 & 28 & 7.1 & 21.7 & 3014.4 & 12628.0 & 15642.4 & 3030.4 & 4676.5 & 7706.9 & 2970.4 & \textbf{800.1} & \textbf{3770.5} & 0.02 & \textbf{2969.4} & 1115.2 & 4084.6 & 0.02 \\
\cmidrule{1-1}\cmidrule{2-5}\cmidrule(lr){6-8}\cmidrule(lr){9-11}\cmidrule(lr){12-15}\cmidrule(lr){16-19}
\multirow{5}{*}{\rotatebox{90}{DWR}} & 10 & 84 & 7.3 & 5.7 & 2219.6 & 11832.1 & 14051.7 & 2158.8 & 7508.4 & 9667.2 & 2098.8 & \textbf{726.3} & \textbf{2825.1} & 0.01 & \textbf{2097.8} & 1015.3 & 3113.1 & 0.02 \\
 & 30 & 84 & 7.3 & 16.0 & 2378.7 & 23776.6 & 26155.2 & 2301.2 & 10246.7 & 12547.9 & 2241.2 & \textbf{694.5} & \textbf{2935.7} & 0.01 & \textbf{2240.2} & 1082.0 & 3322.2 & 0.02 \\
 & 50 & 84 & 7.3 & 26.2 & 2539.8 & 16869.6 & 19409.4 & 2444.1 & 10763.6 & 13207.7 & 2384.1 & \textbf{692.0} & \textbf{3076.1} & 0.01 & \textbf{2383.1} & 1096.5 & 3479.6 & 0.02 \\
 & 70 & 84 & 7.3 & 36.9 & 2707.5 & 15079.9 & 17787.4 & 2592.2 & 12722.2 & 15314.4 & 2532.2 & \textbf{668.6} & \textbf{3200.8} & 0.01 & \textbf{2531.2} & 1108.4 & 3639.5 & 0.02 \\
 & 100 & 28 & 7.3 & 51.9 & 2945.4 & 24986.2 & 27931.5 & 2801.3 & 10169.8 & 12971.0 & 2741.3 & \textbf{705.9} & \textbf{3447.2} & 0.01 & \textbf{2740.3} & 1034.8 & 3775.1 & 0.02 \\
\cmidrule{1-1}\cmidrule{2-5}\cmidrule(lr){6-8}\cmidrule(lr){9-11}\cmidrule(lr){12-15}\cmidrule(lr){16-19}
\multirow{5}{*}{\rotatebox{90}{\makecell{ALL\\DOMAINS}}} & 10 & 429 & 14.5 & 2.7 & 2135.9 & 13656.4 & 15792.3 & \textbf{2003.3} & 6824.3 & 8827.6 & 2031.0 & \textbf{956.0} & \textbf{2987.0} & 0.01 & 2030.0 & 1216.4 & 3246.4 & 0.02 \\
 & 30 & 429 & 14.5 & 7.4 & 2202.2 & 16258.8 & 18460.9 & 2155.0 & 10301.8 & 12456.7 & 2089.8 & \textbf{918.5} & \textbf{3008.3} & 0.01 & \textbf{2088.8} & 1220.2 & 3309.0 & 0.02 \\
 & 50 & 429 & 14.5 & 11.8 & 2283.6 & 18912.6 & 21196.2 & 2211.3 & 10626.4 & 12837.6 & 2146.1 & \textbf{903.5} & \textbf{3049.6} & 0.01 & \textbf{2145.1} & 1241.6 & 3386.6 & 0.02 \\
 & 70 & 429 & 14.5 & 16.8 & 2354.5 & 15529.1 & 17883.7 & 2273.6 & 11220.5 & 13494.0 & 2208.4 & \textbf{914.2} & \textbf{3122.6} & 0.01 & \textbf{2207.4} & 1232.5 & 3439.9 & 0.02 \\
 & 100 & 163 & 14.1 & 21.3 & 2395.4 & 18072.7 & 20468.0 & 2331.3 & 7836.0 & 10167.4 & 2266.3 & \textbf{882.9} & \textbf{3149.1} & 0.01 & \textbf{2265.3} & 1127.8 & 3393.0 & 0.02 \\
\bottomrule
\end{tabular}
}
\end{table*}

\section{Reasoning Excerpts}
\label{app:reasoning}

\begin{figure}[h!p]
\begin{lstlisting}[style=prompt]
The single observed action is (PICK-UP O), which means that the agent's immediate intent was to grasp the block 'O' from the table. This operation directly affects the precondition (ONTABLE O), removing it and causing the holding condition for 'O' to be true while reducing the hand being empty status to false. Evaluating the hypotheses, the most likely goal should involve subsequent actions that make sense in the wake of the current action, in this case removing the restrictions upon (CLEAR M),(ONTABLE W),(ON M O),(ON O W) needed for further stacking or ordering, particularly those involving 'O'. The hypothesis (CLEAR M),(ONTABLE W),(ON M O),(ON O W) suggests that the action of picking up 'O' from the table is a first logical step, supporting future moves to potentially create this arrangement, including placing 'M' on 'O' or 'W', indicating it's a highly plausible sequence of intent, hence scoring highest.
\end{lstlisting}
\caption{\gptfour's reasoning for a problem with 10\% observability in the \textit{Blocks World} domain. The model correctly identifies the most likely goal hypothesis based on the single observed action.}
\label{fig:gpt-4o-correct-reasoning-example-1}
\end{figure}

\begin{figure}[h!p]
\begin{lstlisting}[style=prompt]
The observed action, (PUT-DOWN D), directly affects the placement of block D. However, none of the goal hypotheses explicitly mention the block D being on the table or in any specific configuration that would uniquely indicate the pursuit of such a goal given only this step so far. Since no unique distinguishing action was observed, we assume equal likelihoods for each goal hypothesis considering there is just one observed action and no specific additional evidence favoring any single hypothesis over others. Therefore, all hypotheses are equally probable under the conditions of the given data.
\end{lstlisting}
\caption{\gptfour's reasoning for a problem with 10\% observability in the \textit{Blocks World} domain. The model reasons that the single observed action does not provide enough evidence to discriminate among the candidate goals, leading it to assign equal likelihoods to all hypotheses. This results in a high spread and low precision, even though the correct intended goal is included in the returned set.}
\label{fig:gpt-4o-correct-reasoning-example-2}
\end{figure}

\begin{figure}[h!p]
\begin{lstlisting}[style=prompt]
The observed action is "PICK-UP P", which indicates that the agent's first move was to pick up block P from its initial position on the table. This move directly affects hypotheses where block P is involved in transitioning from its initial state. Hypotheses that reasonably involve moving P or requiring P as a part of the stack tend to have higher scores. The action does not strongly discriminate the specific arrangement of blocks post-initiation but highlights the immediate importance of P. Hypotheses that do not involve P or the area directly around P were scored lower. Consequently, the goals most compatible with this initial action involve P in a stack or clear position, suggesting various sequential possibilities, indicated by the variety of top-scoring hypotheses. Normalizing these based on relativity to P's starting state resulted in the final scores.
\end{lstlisting}
\caption{\gptfour's reasoning for a problem with 10\% observability in the \textit{Blocks World} domain. The model does narrow down to $8$ selected hypotheses from $21$ available. However, it is wrong in this case, as the true goal does not prominently involve block P.}
\label{fig:gpt-4o-wrong-reasoning-example-1}
\end{figure}

\begin{figure}[h!p]
\begin{lstlisting}[style=prompt]
The initial state and the observed action suggest that the agent's first move was to stack block O onto block R. This action is executed from a state where O is clear and on the table, and R is also clear. After performing STACK O R, both blocks O and R will no longer be clear due to being part of a stack, and the agent's hand will be empty again.

From the goal hypotheses, any goal that includes (ON O R) or has an arrangement that could have O directly on R could possibly have been targeted by this action. However, as the only observed action is STACK O R, it doesn't provide enough information to distinctly prefer one goal over another as it could be a step towards many different configurations described in the hypotheses.

Thus, every goal hypothesis receives an equal likelihood score, reflecting the uncertainty given the lack of more distinctive actions or evidence to suggest a particular goal over others.
\end{lstlisting}
\caption{\gptfour's reasoning for a problem with 10\% observability in the \textit{Blocks World} domain. The model reasons about a possible way to select a smaller set of hypotheses but ends up selecting all of them.}
\label{fig:gpt-4o-correct-but-wrong-reasoning-example-1}
\end{figure}

\begin{figure}[h!p]
\begin{lstlisting}[style=prompt]
The observed actions are (PICK-UP A), (STACK A S), and (STACK C A). The sequence of actions begins with picking up block A and placing it on block S, followed by placing block C on top of block A. This indicates a build-up with C at the top of the stack involving A and S. Among the goal hypotheses, (CLEAR C), (ONTABLE H), (ON C A), (ON A S), (ON S H) directly aligns with this sequence of observed actions since it results in a stack of C on A, and A on S, consistent with the observed agent behavior. While there may be other hypotheses that could eventually end up with A, S, and C in some form, none align specifically with the stacking pattern of the observed actions as closely as this hypothesis. Other hypotheses have lower scores due to their lack of direct correlation with the specific order and purpose of the observed stack operations.
\end{lstlisting}
\caption{\gptfour's reasoning for a problem with 30\% observability in the \textit{Blocks World} domain. The model correctly identifies the most likely goal hypothesis based on the observed sequence of actions, which clearly indicates a stacking pattern involving blocks A, S, and C.}
\label{fig:gpt-4o-correct-reasoning-example-3}
\end{figure}

\begin{figure}[h!p]
\begin{lstlisting}[style=prompt]
The observed actions indicate an intent to rearrange the blocks in a manner that correlates most closely with two of the hypotheses. The actions include several unstacking and stacking operations where blocks are rearranged to achieve a very specific configuration pattern. Notably:

1. The initial unstacking of block B from G and subsequent stacking onto J suggests an effort to create a sub-structure involving (ON A J), reflective of moving towards a goal involving stacks initiated by A/J.
2. Actions such as stacking K on F and G suggest an intent to form a sequence that can end with (ON F E), (ON E G), corresponding directly with hypothesis patterns where F is involved in the base transitions of the structure.
3. Moreover, the repeated adjustments involving D and J imply significant reordering involving these blocks, in line with the corresponding hypotheses.
4. Finally, observed tasks like stacking A on J directly align with a sequence that follows the thematic transition in two of the hypothesized stack orders, indicating a prioritized sequence in these constructs.

The two hypotheses that arise as most likely are those with a focus on initial D/J/A to H projects transitioning into wider stack compositions, represented in the final, observed movements and configurations occurring drastically at the functional goal end.
\end{lstlisting}
\caption{\gptfour's reasoning for a problem with 70\% observability in the \textit{Blocks World} domain. The model fails to reason on a more complex problem with a larger number of objects involved and narrows down to 2 wrong hypotheses.}
\label{fig:gpt-4o-wrong-reasoning-example-3}
\end{figure}

\begin{figure}[h!p]
\begin{lstlisting}[style=prompt]
To evaluate which goal hypothesis is most likely, we examine how the sequence of actions aligns with each goal state's required configuration. The sequence of actions observed is such that the final goal state must have block T on block H, block A on block W, and block R on the table.

1. The UNSTACK T W and STACK T M operations suggest T is involved in the final configuration.
2. Subsequently, UNSTACK R A, PUT-DOWN R, PICK-UP A, and STACK A W indicate that R must be on the table while A is on W.
3. Finally, STACK T H places T on H, making H a supporting block, implying H must support other blocks.

Given the executed actions, the only hypothesis producing such a state is (CLEAR M),(ONTABLE W),(ON M O),(ON O W). This configuration would have M and O stacked together, with M clear and W clear, and the rest of the blocks not violating any constraints from actions. The entire configuration needed to execute the observed actions fits perfectly with this particular goal hypothesis. Hence, this goal hypothesis is assigned a likelihood score of 1.0 and is selected as the sole most likely goal.
\end{lstlisting}
\caption{\gptfour's reasoning for a problem with 70\% observability in the \textit{Blocks World} domain. The model wrongly selects the goal hypothesis (CLEAR M),(ONTABLE W),(ON M O),(ON O W) as the most likely one, based on the observed sequence of actions. However, the true goal does not prominently involve block M or O, leading to an incorrect prediction based on a hallucination.}
\label{fig:gpt-4o-wrong-reasoning-example-2}
\end{figure}

\begin{figure}[h!p]
\begin{lstlisting}[style=prompt]
The observed actions (UNSTACK R P), (PICK-UP P), and (PICK-UP R) indicate a focus on manipulating blocks R and P. The unstacking of R from P and then picking up these blocks suggests that the goal involves clearing the top of block R. Among the goal hypotheses, the goal (CLEAR R),(ONTABLE W),(ON R A),(ON A W) implies such a manipulation, as it requires R to be moved, possibly involving unstacking operations and ensuring R is clear. This hypothesis aligns with the observed actions, which seemed to clear R and reposition it, making it the most likely goal. Other hypotheses receive lower scores due to less alignment with the observed actions, either involving different block arrangements or predicting different sequencing of block manipulations.
\end{lstlisting}
\caption{\gptfour's reasoning for a problem with 30\% observability in the \textit{Blocks World} domain. The model fails to correctly identify the most likely goal due to limited observability even though there were hypotheses involving both R and P.}
\label{fig:gpt-4o-wrong-reasoning-example-4}
\end{figure}

\begin{figure}[h!p]
\begin{lstlisting}[style=prompt]
The observed action is a single STACK D E. To be applicable, the agent must have been holding block D and have block E clear. This implies that prior to the action D was on some block and had been unstacked, while E was clear. After the action, the resulting facts are: CLEAR D, ON D E, HANDEMPTY, and NOT CLEAR E. For each goal hypothesis we counted how many of its required facts are satisfied in this resulting state. The counts were:

Hypothesis 19: 3 facts satisfied (CLEAR W, ONTABLE E, ON D E)
All other hypotheses satisfied at most 2 facts, some only 1 or 0.
Because the action directly contributes to satisfying ON D E (present only in hypotheses 12 and 19) and also contributes to CLEAR W and ONTABLE E (present in some), hypothesis 19 attains the highest number of satisfied facts. We use a simple proportional likelihood model: the likelihood of each hypothesis is the number of satisfied facts divided by the total number of satisfied facts across all hypotheses, ensuring the scores sum to 1. This gives hypothesis 19 a normalized score of 3/35 = 0.0857, which is higher than any other hypothesis. Hence hypothesis 19 is the most likely goal.
\end{lstlisting}
\caption{\gptoss's reasoning for a problem with 10\% observability in the \textit{Blocks World} domain. The model correctly applies its fact-matching strategy to identify the most likely goal.}
\label{fig:gpt-oss-correct-reasoning-example-1}
\end{figure}

\begin{figure}[h!p]
\begin{lstlisting}[style=prompt]
I evaluated each goal hypothesis by checking how many of its facts are present in the state after executing the observed actions (ignoring precondition conflicts). For each goal I computed a match ratio = (# of matched facts)/(total facts in the hypothesis). These ratios were then normalized so that all scores sum to 1. The goal with the highest ratio was the one where all its required facts were satisfied in the final state: (CLEAR A),(ONTABLE H),(ON A S),(ON S H). Since it achieved a perfect match (4/4) this hypothesis received the highest likelihood score of 0.1091, making it the single most likely goal among those provided. The remaining hypotheses received lower scores proportionally to how many of their goal facts matched the final state.
\end{lstlisting}
\caption{\gptoss's reasoning for a problem with 10\% observability in the \textit{Blocks World} domain. The model fails in identifying the correct goal by considering an intermediate state instead of the final one, leading it to select an incorrect hypothesis.}
\label{fig:gpt-oss-wrong-reasoning-example-1}
\end{figure}

\begin{figure}[h!p]
\begin{lstlisting}[style=prompt]
All 21 goal hypotheses are treated as equally admissible because the observed action sequence is inconsistent with the domain pre-conditions, making it impossible to determine which goal the agent actually pursues. Since none of the goals can be ruled out based on the observations, each hypothesis receives an equal likelihood of 1/21 = 0.047619, and therefore all goals tie as the most likely.
\end{lstlisting}
\caption{\gptoss's reasoning for a problem with 30\% observability in the \textit{Blocks World} domain. The model assigns a uniform likelihood to all hypotheses due to the observed action sequence being inconsistent with the domain preconditions, resulting in a failure to discriminate among the candidate goals.}
\label{fig:gpt-oss-wrong-reasoning-example-2}
\end{figure}

\begin{figure}[h!p]
\begin{lstlisting}[style=prompt]
Initial state: A is on C, D is on A, R is on P, and O, E, W are on the table and clear; hand is empty.

Observed actions:

1. UNSTACK A C - applicable, agent holds A, C becomes clear.
2. PICK-UP P - not applicable immediately after action 1 in strict Blocks World semantics, because after unstacking A the hand is not empty.

Still, the two key structural intentions revealed are:

- the agent removes A from C, freeing A from below;
- the agent then manipulates P and finally

3. STACK P A - which places P on A.

The final observed relation ON P A is highly diagnostic. Among all hypotheses, only hypothesis 20 contains (ON P A). It also requires (CLEAR P) and (ONTABLE W) and (ON A W). After stacking P on A, P is indeed clear, matching that goal component. This makes hypothesis 20 uniquely aligned with the observation.
\end{lstlisting}
\caption{\gptfive's reasoning for a problem with 30\% observability in the \textit{Blocks World} domain. The model correctly infers the agent's goal despite the observed action sequence violating the domain preconditions, demonstrating a more robust reasoning process compared to \gptoss.}
\label{fig:gpt-5-4-correct-reasoning-example-1}
\end{figure}

\begin{figure}[h!p]
\begin{lstlisting}[style=prompt]
Applying the observed actions from the initial state yields the following final tower structure over the relevant blocks: (ON J G), (ON H J), (ON E H), (ON A E), (ON I A), (ON F I), (ON D F), (ON B D), (ON C B), with G on the table. This exactly corresponds to the chain C-B-D-F-I-A-E-H-J-G when read from top to bottom.

The observations strongly show the agent systematically dismantling the initial towers and rebuilding a single tower... Several actions are redundant undo/redo actions, such as stacking and unstacking e on g, j on e, h on j, and d on f, but they do not change the final state reached after the full observation sequence. The model concludes uniquely that hypothesis 5 describes this final structure.
\end{lstlisting}
\caption{\gptfive's reasoning for a problem with full observability in the \textit{Blocks World} domain. The model fails to identify the correct goal, missing the complete ordering sequence by switching one block out of place, showcasing its difficulty in handling complex problems with many objects and long observation sequences, similar to \gptoss.}
\label{fig:gpt-5-4-wrong-reasoning-example-1}
\end{figure}

\begin{figure}[h!p]
\begin{lstlisting}[style=prompt]
Initial state: U, T, R, K, C are clear and on the table; S is on A, A is on H, and H is on the table; hand is empty.

Observed actions: 1. PICK-UP U, 2. STACK U T, 3. PICK-UP C, 4. STACK C U

These actions build the tower C on U on T, yielding a state where ON U T and ON C U are true, T is on the table, and C is clear.

The observations match hypothesis 8 exactly: (CLEAR C),(ONTABLE T),(ON C U),(ON U T). After the four observed actions, every fact in that goal already holds, so this is a very strong candidate.

Hypothesis 20 is also strongly supported because it contains the achieved substructure (CLEAR C),(ONTABLE T),(ON C U), but requires extending the tower to ON U R and ON R T. This is compatible with the observation pattern: the agent may be assembling part of a larger final tower from the top downward/subtower-first...

Thus only hypotheses 8 and 20 receive nonzero probability. Hypothesis 20 is assigned the highest likelihood because the observations look like a purposeful partial construction toward a larger stack rather than necessarily stopping immediately once hypothesis 8 is reached.
\end{lstlisting}
\caption{\gptfive's reasoning for a problem with full observability in the \textit{Blocks World} domain. The model fails to identify the correct goal, assuming that the observations are on the path to reach the selected hypotheses instead of considering one that is already achieved.}
\label{fig:gpt-5-4-wrong-reasoning-example-2}
\end{figure}

\begin{figure}[h!p]
\begin{lstlisting}[style=prompt]
Initial state: O, E, W are clear and on the table; R is on P and clear; D is on A, A is on C, and C is on the table; hand is empty.

Observed action 1 is (STACK O W). For this to be applicable, the agent must already be holding O and W must be clear. W is clear initially, so the observation strongly indicates that immediately before the observed trace the agent had picked up O. After this action, an achieved relation is (ON O W). This fact appears only in hypotheses 5, 6, 14, and 16, so those goals are supported by the first observation. All other hypotheses are inconsistent with the only positive structural fact created by the first observed action.

Observed action 2 is (UNSTACK R P). This action is applicable in the initial state after action 1, because R is on P, clear, and the hand is empty after stacking O on W. Its effect removes (ON R P), makes the agent hold R, and makes P clear. This is useful when the goal needs R moved elsewhere, or when P must become clear.

Among the goals supported by action 1:
- Hypothesis 5: needs (ON R O) and (ON O W). Since O is already on W, unstacking R from P is a sensible next step toward stacking R on O.
- Hypothesis 6: needs (ON R O) and (ON O W). Same support as hypothesis 5.
- Hypothesis 16: needs (ON C O), (ON O W), (ON W E), (ON E R). This is a longer tower involving O on W, but unstacking R from P can also be relevant because R is the intended base for E.
- Hypothesis 14: needs (ON P O), (ON O W), (ON W E), (ON E R). Here the second action is especially informative: unstacking R from P directly clears P, which is necessary if P is later to be stacked onto O. It also starts relocating R so E can be placed on R. Thus both observations align well with this hypothesis.

There is one additional hypothesis strongly supported by the second observation:
- Hypothesis 20: (CLEAR P),(ONTABLE W),(ON P A),(ON A W). Unstacking R from P achieves (CLEAR P) directly, which is one of the goal facts. Also, to build A on W and P on A, W must remain on the table and be clear enough to receive A; the earlier stack O on W is not itself a goal fact, but it may be an intermediate step that would later need to be undone, making this somewhat less cleanly aligned than the O-on-W goals. Still, the explicit achievement of (CLEAR P) by the second observed action makes this hypothesis very plausible.

Comparing supported hypotheses, hypothesis 20 is the strongest because the second observed action directly achieves one of its required goal facts, whereas for hypotheses 5, 6, 14, 16, and 21 the actions are mostly preparatory except that action 1 achieves (ON O W). Among those, hypotheses 5, 6, 16, and 21 receive some support, but less than hypothesis 20. The remaining hypotheses are given zero because they are not supported by the observed key effects and often conflict with them, especially the creation of (ON O W) or the use of R and P in the second action.

The scores are normalized to sum to 1, with the highest likelihood assigned uniquely to hypothesis 20.
\end{lstlisting}
\caption{\gptfive's reasoning for a problem with $30\%$ observability in the \textit{Blocks World} domain. The model fails to identify the correct goal, overcommitting to a single hypothesis with lacking information.}
\label{fig:gpt-5-4-wrong-reasoning-example-3}
\end{figure}

\begin{figure}[h!p]
\begin{lstlisting}[style=prompt]
The sequence is invalid. After (pick-up E), the robot's hand holds E, so handempty is false. The action (pick-up C) requires handempty to be true. Therefore, you cannot pick up C while holding E. This makes the sequence impossible to execute as described.
\end{lstlisting}
\caption{\qwen's reasoning for a problem with $10\%$ observability in the \textit{Blocks World} domain. The model identifies the precondition violation but refuses to continue from there.}
\label{fig:qwen3-5-wrong-reasoning-example-1}
\end{figure}

\begin{figure}[h!p]
\begin{lstlisting}[style=prompt]
Hypothesis 1: `(ONTABLE B)(ON S A)(CLEAR D)(ONTABLE D)(ON S T)`

Based on the provided text, there is significant confusion regarding the status of `(ON S T)` in Hypothesis 1 and whether `A` is `T`. The text "Wait `(ON S T)` is in Hypothesis 1." suggests that Hypothesis 1 contains `(ON S T)`. Additionally, the text "Wait `(ONTABLE D)` is in Hypothesis 1." and "If both `(ONTABLE D)` and `(ON S T)` are in Hyp 1." implies that both are likely true. The hypothesis `(ONTABLE B)(ON S A)(CLEAR D)(ONTABLE D)(ON S T)` contains both `(ONTABLE D)` and `(ON S T)`. The fact that the text explicitly states "Wait `(ON S T)` is in Hypothesis 1" and that this is a statement of fact within the provided logic, supports the inclusion of `(ON S T)` in Hypothesis 1. The text "If `A` is `T`" might be a red herring or a condition that is assumed. Given the conflicting statements, the hypothesis that contains `(ON S T)` and `(ONTABLE D)` is Hypothesis 1, as described in the text. Therefore, the most consistent hypothesis given the explicit "Wait" statements is Hypothesis 1.

Hypothesis 1: `(ONTABLE B)(ON S A)(CLEAR D)(ONTABLE D)(ON S T)`
\end{lstlisting}
\caption{\qwen's reasoning for a problem with $10\%$ observability in the \textit{Blocks World} domain. The model generates unsound reasoning, generates output outside the required format, and selects a hypothesis that was not even among the provided goal hypotheses, showcasing a complete failure in understanding the task and the provided information.}
\label{fig:qwen3-5-wrong-reasoning-example-2}
\end{figure}

\begin{figure}[h!p]
\begin{lstlisting}[style=prompt]
You put x on table.
\end{lstlisting}
\caption{\qwen's reasoning for a problem with $30\%$ observability in the \textit{Blocks World} domain. The model fails completely and generates a nonsensical response.}
\label{fig:qwen3-5-wrong-reasoning-example-3}
\end{figure}

\begin{figure}[h!p]
\begin{lstlisting}[style=prompt]
[:answer A]
\end{lstlisting}
\caption{\qwen's reasoning for a problem with $30\%$ observability in the \textit{Blocks World} domain. The model fails completely and generates a nonsensical response.}
\label{fig:qwen3-5-wrong-reasoning-example-4}
\end{figure}

\begin{figure}[h!p]
\begin{lstlisting}[style=prompt]
After the first two actions (UNSTACK R P and PUT-DOWN R), the block R is placed on the table, and P becomes clear. The next two actions (PICK-UP E and STACK E R) result in E being stacked on R. This establishes the state where (ON E R) holds and R is on the table (ONTABLE R).

Among the goal hypotheses, only Hypothesis 14 and Hypothesis 16 contain the facts (ON E R) and (ONTABLE R). Hypothesis 16 requires (CLEAR C) to be true and (ON C O) to hold. However, block C is currently under block A (since (ON A C) is in the initial state and A has not been moved), implying (CLEAR C) is false. Thus, Hypothesis 16 is inconsistent with the current state and requires moving C (supported by A), which is not implied by the actions. Hypothesis 14 requires (CLEAR P) and (ON P O). (CLEAR P) is true after action 1. Since Hypothesis 14 is the only hypothesis compatible with the derived stack E on R, it is the most likely goal.
\end{lstlisting}
\caption{\qwen's reasoning for a problem with $50\%$ observability in the \textit{Blocks World} domain. The model correctly identifies the most likely goal hypothesis based on the observed actions and state changes.}
\label{fig:qwen3-5-correct-reasoning-example-1}
\end{figure}

\clearpage
\newpage
\section*{NeurIPS Paper Checklist}

\begin{enumerate}

\item {\bf Claims}
    \item[] Question: Do the main claims made in the abstract and introduction accurately reflect the paper's contributions and scope?
    \item[] Answer: \answerYes{} 
    \item[] Justification: The claims made in the abstract and introduction are backed by the results in the paper, in Sections~\ref{sec:results} and~\ref{sec:discussion}.

\item {\bf Limitations}
    \item[] Question: Does the paper discuss the limitations of the work performed by the authors?
    \item[] Answer: \answerYes{} 
    \item[] Justification: The paper discusses the limitations of the work in the conclusion section (Section~\ref{sec:conclusion}).

\item {\bf Theory assumptions and proofs}
    \item[] Question: For each theoretical result, does the paper provide the full set of assumptions and a complete (and correct) proof?
    \item[] Answer: \answerNA{} 
    \item[] Justification: The paper does not include theoretical results.

    \item {\bf Experimental result reproducibility}
    \item[] Question: Does the paper fully disclose all the information needed to reproduce the main experimental results of the paper to the extent that it affects the main claims and/or conclusions of the paper (regardless of whether the code and data are provided or not)?
    \item[] Answer: \answerYes{} 
    \item[] Justification: The paper provides detailed instructions and information necessary to reproduce the main experimental results. The landmark-based method used as baseline is available in public repository, and the prompt and models used in the LLM experiments are fully disclaimed.

\item {\bf Open access to data and code}
    \item[] Question: Does the paper provide open access to the data and code, with sufficient instructions to faithfully reproduce the main experimental results, as described in supplemental material?
    \item[] Answer: \answerYes{} 
    \item[] Justification: The paper provides detailed instructions and information necessary to reproduce the main experimental results. Moreover, the code is made available as compressed file along the submission. The code for the landmark-based baseline approach is available in public repository.

\item {\bf Experimental setting/details}
    \item[] Question: Does the paper specify all the training and test details (e.g., data splits, hyperparameters, how they were chosen, type of optimizer) necessary to understand the results?
    \item[] Answer: \answerYes{} 
    \item[] Justification: Since no training is performed, only the usage of off-the-shelf LLMs, the prompt template fully disclosed and the benchmark dataset disclosed and available in public repository, yes.

\item {\bf Experiment statistical significance}
    \item[] Question: Does the paper report error bars suitably and correctly defined or other appropriate information about the statistical significance of the experiments?
    \item[] Answer: \answerNo{} 
    \item[] Justification: The paper does not include multiple runs of the same experiment, so error bars are not applicable.

\item {\bf Experiments compute resources}
    \item[] Question: For each experiment, does the paper provide sufficient information on the computer resources (type of compute workers, memory, time of execution) needed to reproduce the experiments?
    \item[] Answer: \answerYes{} 
    \item[] Justification: The paper provides detailed information about the compute resources used for the experiments, including the type of CPU and GPU, memory, and operating system.
    
\item {\bf Code of ethics}
    \item[] Question: Does the research conducted in the paper conform, in every respect, with the NeurIPS Code of Ethics \url{https://neurips.cc/public/EthicsGuidelines}?
    \item[] Answer: \answerYes{} 
    \item[] Justification: the work conducted in this research has in no form violated the NeurIPS Code of Ethics.

\item {\bf Broader impacts}
    \item[] Question: Does the paper discuss both potential positive societal impacts and negative societal impacts of the work performed?
    \item[] Answer: \answerNA{} 
    \item[] Justification: No societal impacts are expected from the work performed in this research.
    
\item {\bf Safeguards}
    \item[] Question: Does the paper describe safeguards that have been put in place for responsible release of data or models that have a high risk for misuse (e.g., pre-trained language models, image generators, or scraped datasets)?
    \item[] Answer: \answerNA{} 
    \item[] Justification: The paper poses no such risks. All data made available to models are benchmarking problems that are already publicly available.

\item {\bf Licenses for existing assets}
    \item[] Question: Are the creators or original owners of assets (e.g., code, data, models), used in the paper, properly credited and are the license and terms of use explicitly mentioned and properly respected?
    \item[] Answer: \answerYes{} 
    \item[] Justification: The paper explicitly mentions the models used and the companies that created them.

\item {\bf New assets}
    \item[] Question: Are new assets introduced in the paper well documented and is the documentation provided alongside the assets?
    \item[] Answer: \answerNA{} 
    \item[] Justification: The paper does not release new assets. It just benchmarks existing models on an existing dataset.

\item {\bf Crowdsourcing and research with human subjects}
    \item[] Question: For crowdsourcing experiments and research with human subjects, does the paper include the full text of instructions given to participants and screenshots, if applicable, as well as details about compensation (if any)? 
    \item[] Answer: \answerNA{} 
    \item[] Justification: The paper does not involve crowdsourcing nor research with human subjects.

\item {\bf Institutional review board (IRB) approvals or equivalent for research with human subjects}
    \item[] Question: Does the paper describe potential risks incurred by study participants, whether such risks were disclosed to the subjects, and whether Institutional Review Board (IRB) approvals (or an equivalent approval/review based on the requirements of your country or institution) were obtained?
    \item[] Answer: \answerNA{} 
    \item[] Justification: The paper does not involve research with human subjects.

\item {\bf Declaration of LLM usage}
    \item[] Question: Does the paper describe the usage of LLMs if it is an important, original, or non-standard component of the core methods in this research? Note that if the LLM is used only for writing, editing, or formatting purposes and does \emph{not} impact the core methodology, scientific rigor, or originality of the research, declaration is not required.
    \item[] Answer: \answerYes{} 
    \item[] Justification: The main goal of the paper is benchmarking LLMs in a specific problem setting and that is disclosed throughout the paper.

\end{enumerate}

\end{document}